\pdfoutput=1
\documentclass[11pt]{article}

\usepackage[preprint]{acl}

\usepackage{times}
\usepackage{latexsym, amssymb}
\usepackage{algorithm}
\usepackage{algpseudocode}
\usepackage{amsmath}
\usepackage{graphicx}
\usepackage{arydshln}
\usepackage{xcolor}
\usepackage[T1]{fontenc}
\usepackage[utf8]{inputenc}

\usepackage{microtype}

\usepackage{inconsolata}
\usepackage{booktabs,subcaption}
\usepackage{multirow}

\title{Collaborative decoding of critical tokens for boosting factuality of large language models}

\author{Lifeng Jin, Baolin Peng, Linfeng Song, Haitao Mi, Ye Tian and Dong Yu \\
Tencent AI Lab, Bellevue, WA \\
\texttt{lifengjin@global.tencent.com} \\}

\begin{document}
\maketitle
\begin{abstract}
The most common training pipeline for large language models includes pretraining, finetuning and aligning phases, with their respective resulting models, such as the pretrained model and the finetuned model. Finetuned and aligned models show improved abilities of instruction following and safe generation, however their abilities to stay factual about the world are impacted by the finetuning process. Furthermore, the common practice of using sampling during generation also increases chances of hallucination. In this work, we introduce a collaborative decoding framework to harness the high factuality within pretrained models through the concept of critical tokens. We first design a critical token classifier to decide which model to use for the next token, and subsequently generates the next token using different decoding strategies. Experiments with different models and datasets show that our decoding framework is able to reduce model hallucination significantly, showcasing the importance of the collaborative decoding framework.
\end{abstract}

\section{Introduction}
\label{sec:intro}
\begin{figure}
    \centering
    \includegraphics[width=0.42\textwidth]{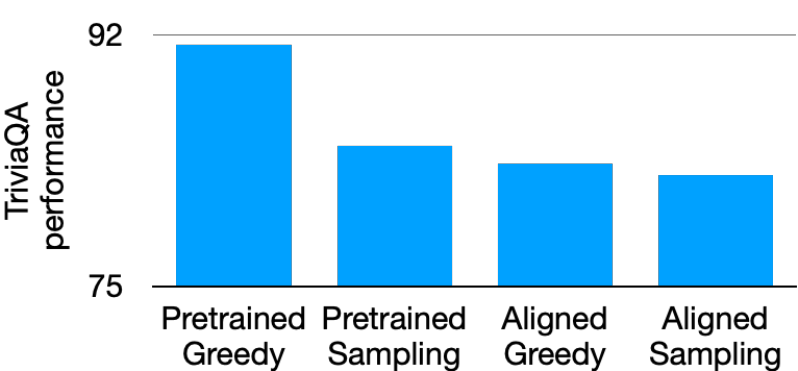}
    \caption{Performance of different LLaMA-2 70B models on TriviaQA. Aligned denotes the Chat models.}
    \label{fig:intro}
\end{figure}

Large language models such as ChatGPT \cite{2022OpenAIchatgpt,2023GPT4Openai} and LLaMA-2 \cite{touvron2023llama2} go through pretraining and aligning stages \cite{ouyang2022training} in order to be helpful and useful AI tools, with the aligned model being the final product for the end users. Although aligning models%
\footnote{We use alignment to broadly refer to the supervised finetuning and human preference alignment processes.} %
to human preferences makes models safer and more helpful, it also costs the model's performance to correctly recall factual knowledge learned during the pretraining stage, increasing its tendency to hallucinate incorrect facts about the world \cite{ouyang2022training,bai2022hhrhlf,chung2022scaling,lin2024mitigating}. In addition, the sampling decoding strategy, commonly used by ordinary users to interact with LLMs, introduces randomness to answers, including places where randomness is not desirable, such as population numbers, birthdays and names of famous people. All of this contributes to the factuality hallucination problem of LLMs \cite{xu2024hallucination}, providing wrong answers to fact-related questions. Figure~\ref{fig:intro} shows performance decline when alignment and sampling are applied to the LLaMA-2 70B models, showing the negative impact of these methods to factuality of a model.\footnote{Details about the evaluation are described in Section \ref{sec:datasets}.}

There are many mitigation strategies already proposed in the literature. External hallucination mitigation strategies include external information retrieval systems for retrieving relevant facts to the questions \cite{shuster2021retrieval,izacard2022few,zhao2023verify} or additional training data targeted for hallucination \cite{li2023inference,zhang2023alleviating}. Internal mitigation strategies, such as logits and representation manipulation methods, do not take advantage of external knowledge sources, but often require dataset-specific tuning~\cite{li2022contrastive,chuang2023dola}. Such tuning limits the generalization abilities of these approaches.

The proposed approach in this work is an internal mitigation strategy which requires no dataset-specific tuning. One motivation for this work is that since both alignment tuning and sampling have a negative impact on model factuality, we can improve it by selectively control the sampling strategy and the prediction model at the token level. Because the tokens affecting model factuality are usually small in number in a generated answer, if such tokens are generated by a more truthful model, such as the pretrained model, and in a more truthful manner, such as with greedy decoding, then the overall factuality of the generated answer is enhanced. In this work, we first define \textbf{the critical tokens} as the tokens whose randomness in predicted distributions is not desirable and potentially harmful to factuality. Then we proposed to use \textbf{a model-based approach} to predict the location of such critical tokens. Finally with the token-level classifier, we connect the pre-alignment-tax model to the main, aligned model and run inference with \textbf{collaborative decoding} with both models.

\begin{figure*}[h]
    \centering
    \includegraphics[width=\textwidth]{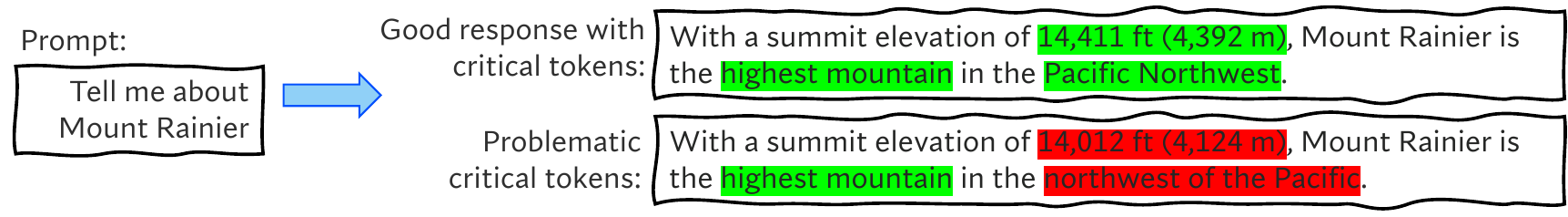}
    \caption{A prompt with a factuality-related question and some model responses. The critical tokens shaded in green in the response include proper names, numbers and facts about the entity, which have low variance tolerance.}
    \label{fig:critical}
\end{figure*}

\begin{figure}[h!]
    \centering
    \includegraphics[width=0.42\textwidth]{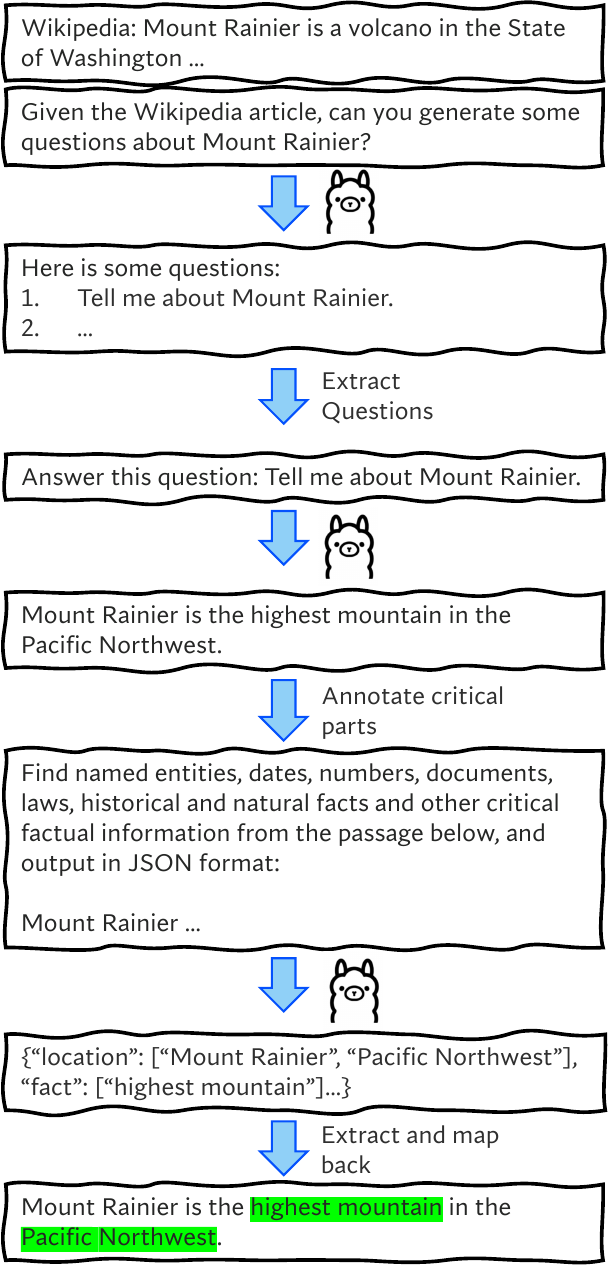}
    \caption{Dataset generation for critical tokens.}
    \label{fig:extraction}
\end{figure}

This approach has several advantages. First, the high factuality of the pretrained models can be exploited with little damage to the instruction following capabilities and generation diversity of the aligned models, due to the small number of critical tokens in the generated answer. Second, being the knowledge model, several settings, such as the sampling method, of the pretrained model are decoupled with the user-controlled inference process of the aligned model, enabling further enhancement of factuality of the final model. Third, our method is not model- or dataset-specific and requires no knowledge resources, allowing easy deployment with good generalization abilities. Finally, the pretrained model as a knowledge model can freely and easily receive knowledge updates through continual training with no impact to the aligned model, making knowledge updates to the whole system simple and effective.

\section{Critical tokens}

\subsection{Critical tokens}
Critical tokens are defined to be a small number of tokens in responses which has a disproportionately large influence on the classification of it in some category. For example, profanity of a sentence is often decided by a couple of profane words. Simply dropping or changing them can turn a profane sentence into a normal one~\cite{deng-etal-2022-cold,zhang-etal-2023-safeconv}.

Factuality of a generated answer can also be influenced by a small number of words and phrases. Such words and phrases include numbers, dates, names of people and locations, and short phrases which contain descriptions of facts, natural phenomena, laws and so on. These words and phrases generally do not tolerate randomness in their distributions of surface or semantic representations given a context, because they need to be repeated nearly exactly to be perceived as correct. One such example can be seen in Figure \ref{fig:critical}, where the height of the mountain and the name of its location do not tolerant disturbances in surface and semantic spaces.

Fortunately, popular LLMs such as Llama 2 \cite{touvron2023llama2} and Mistral \cite{jiang2023mistral} are able to locate such critical tokens by themselves, because frequently such tokens are named entities, numbers and so on. This allows us to freely generate large quantities of data for training critical-token predictors. 
Second, the designation of critical tokens in an answer is independent from the correctness of the answer, therefore we are able to use any answer for critical token annotation. However, answers to fact-related questions may contain more critical tokens compared to other requests such as poetry generation or lyrics analysis, therefore we use fact-related question and answer pairs for automatic annotation.
Furthermore, the critical tokens are usually not sensitive to generation style, therefore the training data generated by one model can be used by other models from the same or different family, provided that the tokenizers are not drastically different.

\subsection{Critical token dataset}

Figure \ref{fig:extraction} shows the process used in this work to generate critical token annotation. A Llama 2 70B Chat model is first asked to generate five factual questions about a given Wikipedia article. It is then asked to generate an answer to each question. We do not examine the correctness of the answers, as discussed above. Then it is asked to extract critical tokens from a given answer in JSON format, and the psuedo-annotation of extracted tokens is mapped back into the original answer to create a sequence prediction dataset. In total, the dataset contains 131180 training instances and 7287 validation and test instances. Each instance contains a sequence of tokens from a question and an answer as the input, and a sequence of \texttt{Yes} and \texttt{No} labels as the output. The total number of \texttt{Yes} label in the dataset is 3 million, which is about 11\% of all labels.

\subsection{Critical token prediction}

With the critical token dataset, a critical token prediction model can be trained to predict whether a particular token is a critical token in a binary classification fashion. We adopt a supervised finetuning approach, training a Llama 2 13B model to output \texttt{Yes} or \texttt{No} given a prefix. The training loss is:
\begin{equation}
    L_{\mathrm{ct}} = - \sum_{n=1}^N \log p(\mathrm{label}_n | \mathrm{prefix}_n),
\end{equation}
where $\mathrm{label} \in \{\text{\texttt{Yes}, \texttt{No}}\}$ and $\mathrm{prefix}$ is constructed with the template 
\begin{equation*}
    \text{\texttt{Question:\ \{prompt\}\ Answer:\ \{response\}}},
\end{equation*}
where the response is an incomplete series of tokens.

\begin{table}[t]
\centering
\begin{tabular}{lllll}
\toprule
   &  \multicolumn{2}{c}{All}  & \multicolumn{2}{c}{Switch} \\ 
   & \texttt{Yes} F1 & Acc.  & \texttt{Yes} F1 & Acc. \\
\midrule
NT      & 78.15  & 81.41 & 70.17 & 76.14  \\
CT      & 80.42  & 83.14 & 79.25 & 82.50 \\
\bottomrule
\end{tabular}
\caption{Experiment results of critical token prediction.}
\label{tab:critical_token}
\end{table}

Performance of the trained model on the test set of the critical token dataset is shown in Table \ref{tab:critical_token}. The overall accuracy of the prediction and the F1 of the \texttt{Yes} prediction are reported on two settings: the first being on the whole set (All), and the second being on the subset where predictions switch from \texttt{No} to \texttt{Yes} (Switch, the beginning of a critical token span). Models trained with two different ways are included here. CT refers to predicting whether the current token is a critical token, and NT refers to predicting the next token being a critical one. The CT model achieves better performance in all settings, which is used in the following experiments.

\section{The Collaborative Decoding Strategy}

Broadly, the Collaborative Decoding Strategy (CDS) is a decoding method where two or more models contribute to the final predicted token distribution, similar to weighted averaging of model ensembles \cite{breiman1996bagging,opitz1999popular}. Given a series of models $\mathcal{M}_1, \mathcal{M}_2 \dots \mathcal{M}_K$, the final CDS distribution of the next token for the incomplete prefix $s_{t-1} = [w_1 \dots w_{t-1}]$ is defined to be
\begin{equation}
    p_{\mathrm{CDS}}(w_t | s_{t-1}) = \sum_{k} \lambda_k p_{\mathcal{M}_k}(w_t | s_{t-1}),
\end{equation}
where $\lambda_k$ is the normalized weight for the $k$th model $\mathcal{M}_k$. With the CDS distribution, an appropriate decoding strategy $f_{w_t}$ can be applied to it to yield the final token:
\begin{equation}
    w_t \sim f_{w_t}(p_{\mathrm{CDS}}(w_t | s_{t-1}))
\end{equation}
The values for $\lambda_k$ and the decoding strategies $f_w$ can be defined freely. Baselines defined in the Section \ref{sec:baselines} vary these definitions for comparison. Motivated by the discussion in Section \ref{sec:intro}, we focus on one definition named Model CDS where two models, the pretrained and the aligned, participate in the collaboration, and $\lambda_k$s are binary values provided by a critical token classifier.

\subsection{Model CDS}

We propose Model CDS in this paper as a way to avoid the negative impact from the alignment tax. The approach utilizes both the high factuality in the pretrained model and high instruction following ability in the aligned model when necessary without any need of finetuning of the main model. The algorithm is shown in Algorithm \ref{alg:modelcds}.
\begin{algorithm}[t]
\caption{The Model CDS algorithm}\label{alg:modelcds}
\begin{algorithmic}
\Procedure{ModelCDS}{The pretrained model $\mathcal{M}_p$, the aligned model $\mathcal{M}_a$, the critical token CT classifier $\mathcal{M}_c$, the initial prefix for the pretrained model, the aligned model and the classifier $s^p,s^a,s^c$}
\State $w^a \sim p_{\mathcal{M}_a} (w|s^a)$
\While{$w^a$ is not a STOP token}
    \State $\hat{s}^c \gets [s^c\ w^a]$ \Comment{Append $w^a$ to $s^c$}
    \State $d \gets \mathcal{M}_{c}(\hat{s}^c)$ \Comment{$d \in \{\texttt{Yes}, \texttt{No}\}$}
    \If{$d = $\texttt{Yes}}
        \State $w \gets \mathrm{argmax}_{w}p_{\mathcal{M}_p}(w|s^p)$
    \Else
        \State $w \gets w^a$
    \EndIf
    \State $s^a \gets [s^a\ w]; s^p \gets [s^p\ w]; s^c \gets [s^c\ w]$
    \If{$w$ is a STOP token}
        \State \Return $s^a$
    \EndIf
    \State $w^a \sim p_{\mathcal{M}_a} (w|s^a)$
\EndWhile
\State \Return $s^a$
\EndProcedure
\end{algorithmic}
\end{algorithm}

The algorithm takes in three different prefixes as its initial input: $s^p, s^a, s^c$. The $s^a$ is the standard prefix for the aligned model, where the system prompt, the history conversations, and the user prompt are all incorporated. This prefix format is usually created in the finetuning process. $s^p$ is the prefix for the pretrained model. This prefix is defined to be a few-shot learning prompt with several example question-answer pairs, which strengthens the instruction following ability of the pretrained model. Lastly, $s^c$ is the prefix used in the critical token prediction training, where it usually contains some instructions about how to make critical token prediction and the incomplete response predicted so far.

The algorithm starts when an assistant response is demanded and ends when the generated token is a \texttt{STOP} token, and it generates the whole response token by token. While in generation, first a next token $w^a$ from the aligned model is generated, assuming the critical token classifier is a CT model. Next, a new classifier prefix $\hat{s}^c$ is constructed by appending the token $w^a$ to the prefix $s^c$ and sent to the classifier for critical token prediction. If the critical token decision $d$ is \texttt{Yes}, it indicates that the generation of the token in this location needs to be handled by the pretrained model. The pretrained model then proceeds to generate $w$ using greedy decoding in the same location as $w^a$. The $w$ is now considered the candidate token at this time step, and is appended to all prefixes. If the decision is \texttt{No}, the token $w^a$ is considered the candidate token and appended to all prefixes. The generation terminates when the candidate token is a \texttt{STOP} token defined by the users, such as an \texttt{EOS} token or other custom \texttt{STOP} tokens.

\section{Experiment}

\subsection{Evaluation datasets}
\label{sec:datasets}
We evaluate our proposed approach against several baselines (Section \ref{sec:baselines}) on three fact-based datasets: TriviaQA \cite{Joshi2017-va}, NaturalQuestions \cite{naturalquestions} and \textsc{FactScore} \cite{min2023factscore} to measure the level of hallucination reduction with different approaches. TriviaQA and NaturalQuestions(NQ) datasets contain 7993 and 3610 evaluation questions respectively, and \textsc{FActScore} contains 500 person names for which different systems are asked to generate biographies. No data from the datasets is used for any kind of training or tuning for our approaches.

In order to simulate real use scenarios with the LLMs, our evaluation target is freeform generation with the aligned models with system prompts, unless specifically specified.
For TriviaQA and NaturalQuestions, because these models tend to give long and comprehensive answers,
the evaluation accuracy metric is `\textit{answer recall}', meaning whether the gold answer strings can be found in the generated natural language answer or not. For \textsc{FActScore}, the evaluation metric is the score provided by the package, which represents the percentage of facts in the generation matching facts extracted from Wikipedia articles as judged by ChatGPT.\footnote{\url{https://github.com/shmsw25/FActScore}}

\begin{table*}[t]
\centering
\small
\begin{tabular}{lrrrrrrrr}
\toprule
   &  \multicolumn{2}{c}{Llama 2 70B}  & \multicolumn{2}{c}{Llama 2 13B} & \multicolumn{2}{c}{Llama 2 7B} & \multicolumn{2}{c}{Mistral 7B}\\ 
   & TriviaQA & NQ  & TriviaQA & NQ   & TriviaQA & NQ & TriviaQA & NQ \\
\midrule
Aligned Sampling           &  82.54  & 46.37 & 73.63 & 38.92 & 66.68 & 35.26 & 76.77 & 41.86 \\
Aligned Greedy$\blacktriangle$         & 83.29     & 46.73  & 75.85 & 40.66 & 69.72 & 36.34 & 77.36 & 43.13 \\
Cont. Decoding \cite{li2022contrastive}$\blacktriangle$ & - & - & 76.81 & 41.74 & 68.27 & 36.97 & 76.09 & 43.79 \\
Cont. Decoding: 13B-7B $\blacktriangle$ & - & - & 77.15 & 41.58 & - & - & - & - \\
ITI \cite{li2023inference}$\blacktriangle$ & - & - & - & - & 48.78 & 22.72 & - & - \\ 
DoLa \cite{chuang2023dola}$\blacktriangle$ & - & - & 77.34 & 41.05 & 70.07 & 37.28 & 77.86 & 41.05 \\
ICD \cite{zhang2023alleviating}$\blacktriangle$ & - & - & 77.00 & 42.68 & 69.07 & \bf 37.39 & 77.35 & 45.48 \\
\midrule
Model Self-CDS         & 84.02    & 46.26 & 76.33 & 40.83 & 66.83 & 33.93 & 77.56  & 44.38 \\
Entropy CDS & 85.52    & 48.64 & 76.50 & 40.19 & 71.10 & 37.23 & 79.03 & 45.35  \\
Soft Mixing CDS & 86.68 & 49.47 & 79.41 & 42.77 & 71.12 & 35.56 & 78.76 & 46.01 \\
% Contrasting CDS & 42.16 & 17.98 & 37.42 & 14.96 & 33.72 & 13.96 & 52.26 & 20.58  \\
% \hdashline
Model CDS (proposed)         & \bf 88.02    & \bf 51.93 & \bf 79.93 & \bf 43.98 & \bf 72.40 & 36.54 &  \bf 80.07 & \bf 47.26 \\
  % + Entropy CDS & RUN & RUN & 74.00 & 39.58 & &  &  80.79 & \bf 47.73 \\
\midrule
\textit{Pretrained Sampling}   & 84.51 & 40.64 & 73.36 & 31.16 & 63.94 & 25.45 & 74.53 & 33.38 \\
\textit{Pretrained Greedy}$\blacktriangle$       & \it 91.37 & \it 52.68 & \it 83.40 & 42.77 & \it 76.80 & \it 38.33 & 76.62 & 35.32  \\
\bottomrule
\end{tabular}

\caption{Experiment results of models from the same pipelines under different decoding strategies. \textbf{Bond fonts} indicate highest performance among the baseline strategies. \textit{Italic fonts} show when the Pretrained Greedy strategy outperforms the other strategies. $\blacktriangle$ indicates greedy decoding in all tables. We are not able to run some baselines with larger models due to availability of necessary models or limited resources.}
\label{tab:main_result}
\end{table*}

\subsection{Baseline strategies}
\label{sec:baselines}
We compare the Model CDS approach to previously published baselines such as Contrastive Decoding \cite{li2022contrastive}, DoLa \cite{chuang2023dola}, Inference Time Intervention \cite{li2023inference} and Induced Contrastive Decoding \cite{zhang2023alleviating} in the all experiments whenever possible. The last three methods require hallucination-related data for training and hyperparameter tuning, where our methods require no in-domain data at all. All of these baselines use greedy decoding, as marked with $\blacktriangle$.%
\footnote{Comparison of greedy and sampling decoding is included in the appendix.Results show sampling decreases performances of these baselines.}

\noindent \textbf{Contrastive Decoding (CD)}: This approach contrasts logits from an expert model to that from an amateur model for more faithful predictions. The pretrained model is used as the amateur model by default when the aligned model with the same size is used as the expert model. If models with different sizes are used, they will be noted as Expert-Amateur, such as 13B-7B. In those settings, both models are aligned models. One important difference between CD and Model CDS is that in CD the expert model is the main leading model which should be both instruction-tuned and with high factuality. In Model CDS, the main leading model is only required to be instruction-tuned, therefore decoupling instruction following and factuality.

\noindent \textbf{Inference Time Intervention (ITI)}: This approach first learns latent vectors with supervised learning on the TruthfulQA dataset \cite{lin2021truthfulqa}. During inference, such vectors are used to bias the hidden states for more truthful representations. We use the baked-in ITI version of Llama 2 7B Chat for comparison.\footnote{\url{https://huggingface.co/likenneth/honest_llama2_chat_7B}}

\noindent \textbf{DoLa}: As a variant of CD, DoLa contrasts predictions from two layers within the same model, where the expert layer is the last layer and the amateur layer is one of the early layers. We use the dynamic layer selection setup from \citet{chuang2023dola} with the top half of even layers as candidates.%
\footnote{\url{https://github.com/voidism/DoLa}}

\noindent \textbf{Induced Contrastive Decoding (ICD)}: Similar to CD, this approach uses a contrastive setup. In this framework, the amateur model is finetuned on hallucinated texts from either HaluEval \cite{li2023halueval} or biographies from Wikipedia. We use the HaluEval-trained amateur model in the QA experiment, and the biography-trained in the \textsc{FActScore} experiment, following \citet{zhang2023alleviating}.\footnote{See \url{https://github.com/HillZhang1999/ICD}}

We also propose various CDS methods to compare in both \textsc{FActScore} and TriviaQA and NQ experiments:%

\noindent \textbf{Model CDS}: This is the proposed strategy where a model is trained to predict critical tokens and assigns different models for prediction of such tokens.

\noindent \textbf{Aligned/Pretrained Sampling and Greedy Decodings}: The aligned and the pretrained model is used either with the sampling strategy with temperature 1.0, or with greedy decoding strategy. These are commonly used strategies in practice. The pretrained model is used with 5 shots to ensure that it generates an answer to a question.

\noindent \textbf{Model Self-CDS}: Under this baseline strategy, the aligned model switches to greedy decoding from sampling when the factual information router decides the current token to be a critical token, so that the effect of random sampling may be mitigated by the aligned model itself at high-entropy tokens. This is similar to High Entropy Word Spotting and Replacement \cite{rawte-etal-2023-troubling}. For Llama 2 models, $\gamma=0.9$; for Mistral models $\gamma=1.3$, which account for around 10\% of the tokens.

\noindent \textbf{Entropy CDS}: Under this baseline strategy, the entropy of the predicted distribution of the next token is used for the routing decision. If the entropy is higher than a threshold $\gamma$, the pretrained model is used with greedy decoding. Entropy thresholds are the same as the Model Self-CDS.

\noindent \textbf{Soft Mixing CDS}: Under this baseline strategy, the probabilities from both the aligned and the pretrained model are mixed through interpolation to form an updated distribution for the critical tokens, and greedy decoding is used to find the next token. The mixing ratio is 0.5.

Models from the Llama 2 family are used in the following experiments with the strategies. Furthermore, Mistral 7B models are also used to investigate the robustness of the proposed method.

\subsection{Experiments: Collaborative Decoding increases model factuality}
\label{sec:experiments}

We conduct two sets of experiments with models of different sizes and from different families to investigate how different decoding strategies help boost factuality of the models.\footnote{Runs with multiple seeds are not possible for all experimental conditions due to the large amount of computation required, so all experiments are run once. Boostrap resampling shows a standard deviation of 0.4 for the TriviaQA dataset, 0.6 for NQ for most sampling strategies. Results of published baselines on \textsc{FActScore} are from \citet{zhang2023alleviating}.}

\subsubsection{Performance boost from the pretrained model in the same pipeline}

In this experiment, we use models from the same LLM pipeline, i.e. the pretrained and the aligned model from the same LLM training sequence, to compare different baseline strategies. For example, Llama 2 7B and Llama 2 7B Chat are two models from the same pipeline. Experiment results can be found in Table \ref{tab:main_result} for TriviaQA and NaturalQuestions and Table \ref{tab:factscore} for \textsc{FActScore}.

\noindent \textbf{TriviaQA and NaturalQuestions}: First, the proposed strategy, Model CDS, shows the strongest performance among all baselines across different model sizes and families, indicating the effectiveness of the approach. With the same sampling temperature as the Aligned Sampling strategy, our approach is able to outperform the common Aligned Sampling by around 6\% for TriviaQA and around 5\% for NaturalQuestion for Llama 2 models. Model CDS also outperforms the previously proposed methods by large margins, indicating its strong ability of enhancing model factuality.
This shows that Model CDS is able to achieve high factuality without resorting to a greedy decoding strategy, which would greatly limit the diversity of generated responses. Since most tokens decoded greedily in Model CDS are the tokens which should not show large degrees of variation, it has little impact to the diversity of the responses while providing high levels of factuality, as shown in Section \ref{sec:diversity}.

Similar substantial performance gains from Model CDS can also be observed with Mistral 7B models, where they outperform Llama 2 13B models. Mistral 7B models are \textbf{out-of-domain models} for our approach, because the critical token classifier is trained on Llama 2 70B model output tokens, and the tokenizers of the two families are similar but different. This supports the assumption that the classification of critical tokens is not sensitive to model families and finetuning data, whenever their tokenizers are sufficiently similar.

Second, results also show that various CDS baselines all help the aligned model achieve higher accuracy, especially Soft Mixing CDS and Entropy CDS. This shows that partially mixing the token level distribution from the pretrained model into the distribution from the aligned model is already beneficial for factuality.

Lastly and surprisingly, Model CDS is able to outperform Pretrained Greedy for Mistral 7B. The Pretrained Greedy performance can be considered ceiling performance of Model CDS, because the factuality of the strategy depends on the factuality of the pretrained model. This can be observed in the Llama 2 models. However, the higher performance of Model CDS compared to Pretrained Greedy in this case indicates that the way that the pretrained model is used in these downstream tasks needs be carefully designed for it to reach full potential. Because the few-shot prompt for the Mistral pretrained model is designed with Llama 2 70B models, the mismatch may have caused Mistral Pretrained Greedy to perform worse than Mistral Aligned Greedy. However, Model CDS is able to guide the pretrained Mistral model to overcome the mismatch and performs higher than both pretrained and aligned models, which reduces the importance of prompt design for few-shot prompting for the pretrained model.

\noindent \textbf{\textsc{FActScore}}: Table \ref{tab:factscore} shows the evaluation results on the \textsc{FActScore} dataset. First, Model CDS performs the best among the baselines except for Aligned Greedy and ICD, showing the consistent performance boost from the proposed method compared to other similar baselines. The high score of ICD shows the advantage of dataset-specific finetuning, which is indeed effective on in-domain data. Finally, the Aligned Greedy method also performs very well, however Model CDS is able to reach similar performance without sacrificing response diversity at all.

\begin{table}[h]
\centering
% \small
\begin{tabular}{lll}
\toprule
 Strategy  & \multicolumn{2}{c}{\textsc{FActScore}} \\ 
 & \% R.  & Score\\
\midrule
A. Sampling    & 42.2 & 58.0 \\
A. Greedy$\blacktriangle$      & 37.5 & \it 63.8 \\
\midrule
CD \cite{li2022contrastive} 13B-7B$\blacktriangle$ & 74.2 & 53.5 \\
CD \cite{li2022contrastive} 70B-7B$\blacktriangle$ & 62.2 & 60.3 \\
\midrule
ITI \cite{li2023inference}$\blacktriangle$ &  41.9 &  62.4 \\
DoLa \cite{chuang2023dola}$\blacktriangle$ &  40.7 &  61.3 \\
$\text{ICD}^*$ \cite{zhang2023alleviating}$\blacktriangle$ & 36.1 & \textbf{66.3} \\
\midrule
M. CDS (proposed) & 38.2 & \bf 63.9 \\
\bottomrule
\end{tabular}

\caption{Experiment results of Llama 2 7B based strategies, as well as 13B and 70B based Contrastive Decoding methods, on the \textsc{FActScore} dataset. The shorten names follow the same order as Table \ref{tab:main_result}. $*$ indicates dataset-specific anti-hallucination training. \% R. means the percentage of return responses which contain biographic information. \textbf{X-Y} means the size of the expert model and the contrastive amateur model.}
\label{tab:factscore}
\end{table}

\subsubsection{Mixing models of different sizes and families}
\begin{table}[]
\centering
\begin{tabular}{lll}
\toprule
& TriviaQA & NQ \\
\midrule
\textit{Same family} \\
\hdashline 
A. Sampling 7B     &  66.68 & 35.26  \\
A. Sampling 13B    &  73.63 & 38.92  \\
A. Sampling 70B    &  82.54 & 46.37  \\ 
CD (\citeauthor{li2022contrastive}) 13B-7B$\blacktriangle$ & 77.15 & 41.58 \\
Model CDS 7B+13B    &  75.60  & 39.89 \\
Model CDS 7B+70B    &  80.43  & 43.77 \\
Model CDS 13B+70B   &  84.89  & 47.76 \\
\midrule
\textit{Different family} \\
\hdashline
Model CDS (7B L.+7B M.)   & 71.50   &  37.65  \\
\bottomrule
\end{tabular}
\caption{Experiment results of models with different sizes from the same pipelines. \textbf{X+Y} in the table shows the size of the aligned model with the size of the pretrained model. The last row shows the setup with a Llama 2 7B Chat as the aligned model, and the Mistral 7B as the pretrained model.}
\label{tab:mixing_result}
\end{table}

In this set of experiments, models of different sizes and families are linked together by Model CDS to investigate the possibility of using a small aligned model as a guide model to retrieve fact-related information from large pretrained models as external knowledge models. This is desirable because the majority number of tokens is generated autoregressively by the small model and treated as input by the large model, reducing the generation time compared to a full generation by the large model while enjoying levels of factuality higher than the small model.

Table \ref{tab:mixing_result} shows the results of mixing models. The Aligned Sampling results are copied from Table \ref{tab:main_result} for easy comparison. Results show that Model CDS is able to utilize the larger pretrained models for retrieving more factual answers. For example, in the setup of Model CDS (7B + 70B), where the aligned model is a 7B Chat model and the pretrained is a 70B model, the performance gain over Aligned Sampling is 13.75 points on TriviaQA and 8.51 points on NaturalQuestions. This is also much higher than Model CDS with 7B models only, gaining 8.03 and 7.23 points respectively. The setup 13B + 70B performs better than 70B aligned models when used by themselves, again showcasing the ability of model collaboration under the framework.

Finally, Model CDS (7B L. + 7B M.) shows results of linking two models from two different families together. It performs similarly to Model CDS with Llama 7B models but lower than using only Mistral models, showing that the cross-family linking can be beneficial but
the style and tokenization mismatch of the two models should be addressed carefully to achieve best results.

\subsection{Analysis and Discussion}

We first examine the diversity of responses generated by Model CDS to investigate how much more deterministic the decoding has become when the critical tokens are generated greedily.
We also examine the effect of the number of shots for the pretrained model in Model CDS for more understanding of the role the shots play in keeping the pretrained model on task compared to the generated response prefix from the aligned model.
Examples of generations are also provided.

\subsubsection{Diversity of responses}
\label{sec:diversity}
\begin{figure}
    \centering
    \includegraphics[width=0.42\textwidth]{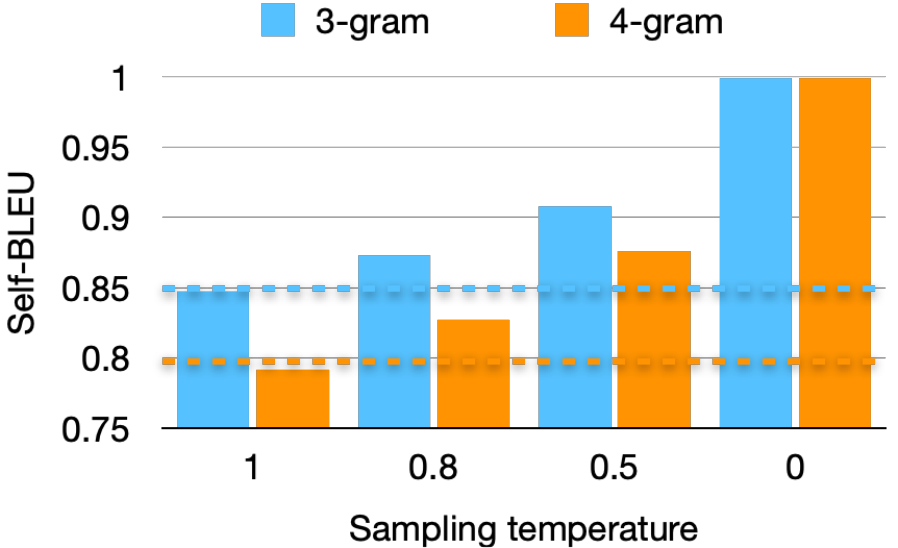}
    \caption{The response diversity of sampling at different temperatures. The dashed lines show the evaluation values for Model CDS. Llama 2 7B models are used.}
    \label{fig:diversity}
\end{figure}

The diversity of the responses from Model CDS is examined by computing the 3-gram and 4-gram self-BLEU scores
\cite{zhu2018texygen}
of 100 sampled biographies of the first 15 entities in the \textsc{FActScore} dataset, and comparing them to the same metrics computed from sampled responses from the aligned model with different temperatures, all with Llama 2 7B models. Figure \ref{fig:diversity} shows that self-BLEU scores of Model CDS, shown as dashed lines, are almost identical to the scores of the Aligned Sampling scores at temperature 1.0. This confirms that Model CDS has very little effect on the diversity of the sampled responses.

\begin{figure}
    \centering
    \includegraphics[width=0.42\textwidth]{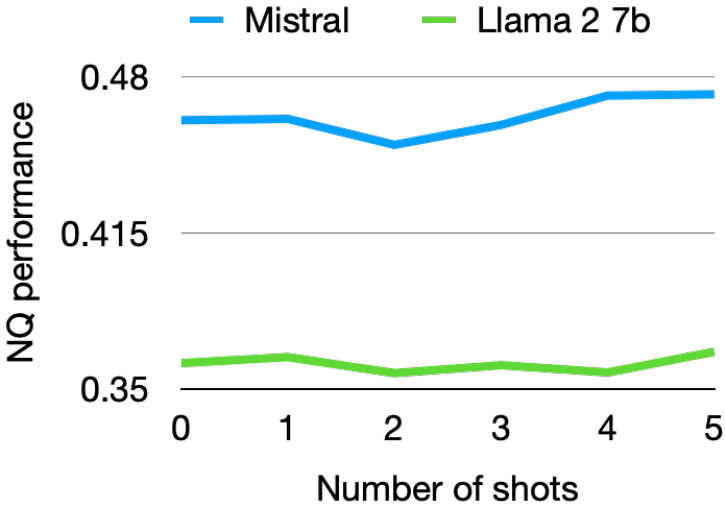}
    \caption{The effect of number of shots for the pretrained model on performance of Natural Questions.}
    \label{fig:shots}
\end{figure}

\subsubsection{Number of shots for the pretrained model}

In this experiment, we vary the number of shots in the prompt for the pretrained model in Model CDS to explore the effect it has on the generation factuality. Figure \ref{fig:shots} shows how performance changes from 0 to 5 shots of question-answer pairs in the prompt. This indicates that the factuality performance is not substantially impacted by the number of shots in the prompt for the pretrained model, although a slight boost in performance can be observed in the results for both models. 

\begin{table*}[t]
\centering
\small
\begin{tabular}{p{0.09\linewidth}p{0.14\linewidth} p{0.69\linewidth}}
\toprule
 System & Approach & Answer \\
\midrule
\multicolumn{3}{l}{\textit{Prompt: when does a wrinkle in time come out in canada?}} \\
\hdashline
\multirow{3}{*}{Mistral}  & Aligned Greedy  &  "A Wrinkle in Time," which is a 2018 American science fiction adventure film, was released in Canada on \colorbox{pink}{March 23, 2018}... \\
    &  Aligned Sampling & The first film adaptation of "A Wrinkle in Time" by Madeleine L'Engle, directed by Ava DuVernay, was released in Canada on \colorbox{pink}{March 23, 2018}... \\
    & Model CDS & "A Wrinkle in Time," a novel written by Madeleine L'Engle, was first published in Canada in 1963, but the release year you might be asking about is likely related to one of the film adaptations. The first film adaptation, "A Wrinkle in Time," was released in theaters in Canada on \colorbox{green}{March 9, 2018}. \\
\midrule
\multicolumn{3}{l}{\textit{Prompt: Which artist has a daughter and two sons with Jane Asher, whom he married in 1981?}} \\
\hdashline
\multirow{3}{*}{Llama2 70B}  & Aligned Greedy  &  The artist who has a daughter and two sons with Jane Asher, whom he married in 1981, is \colorbox{pink}{Paul McCartney}. \\
    &  Aligned Sampling & The answer is \colorbox{pink}{Paul McCartney}. Paul McCartney married Jane Asher in 1981, and they have two sons, Sean and James, and a daughter, Beatrice. \\
    & Model CDS & The artist who has a daughter and two sons with Jane Asher, whom he married in 1981, is \colorbox{green}{Gerald Scarfe}... \\
\bottomrule
\end{tabular}
\caption{Generated examples with different strategies and models.}
\label{tab:examples}
\end{table*}

\subsubsection{Qualitative analysis of generated responses}

Facts stored in model parameters and damaged in the alignment process are able to be recovered with Model CDS. This is observed frequently when looking at generated responses from different strategies, shown in Table \ref{tab:examples}. For both questions, the aligned models provide the same wrong answers regardless of decoding strategies. However, with Model CDS, the correct answer is retrieved from the pretrained model.

\section{Related work}

Hallucination has been in intense focus after the advent of LLMs, with numerous attempts to categorize and evaluate it \cite{yao2023llm,huang2023survey,chang2023survey,shi2024thorough}, with some specifically on factuality \cite{wang2023survey,min2023factscore,li2024dawn}. Methods to change prompting \cite{yu2023chain,dhuliawala2023chain}, finetuning \cite{li2023inference,tian2023fine} and information retrieval \cite{lewis2020retrieval,ma2023query,izacard2022few,gao2024retrievalaugmented,zhao2023verify} have shown great results in combating the issue.

Decoding as a way to address the issue of factuality \cite{shi2024thorough} has its own advantages, because it generally tries to manipulate model predictions for higher factuality without extra training or knowledge sources. Methods such as Contrastive Decoding \cite{li2022contrastive,o2023contrastive} and its variants \cite{chuang2023dola,zhang2023alleviating} have shown good performances on different datasets \cite{min2023factscore,lin2021truthfulqa}.
However, these methods all require hyperparameter tuning on specific datasets, which makes them difficult to scale up to online systems.

\section{Conclusion}

In this work, we propose a decoding strategy framework called Collaborative Decoding Strategy (CDS), where models work collaboratively to jointly decode and generate. We also propose the idea of critical tokens, which are tokens that do not generally tolerate variation under some evaluation criteria. The Model CDS, where the collaboration is decided by a critical token classifier, is proposed to exploit high factuality in external knowledge models and enjoy the instruction following capabilities of the aligned model at the same time. Experiments show that with a pretrained model as an external knowledge model, factuality of the aligned model is enhanced significantly with very little negative impact to generation diversity.

\clearpage

\section*{Limitations}

This work explores how models at different stages in the LLM training pipeline can collaborate together for high factuality. The proposed approaches utilize no external knowledge, which effectively sets the factuality of the pretrained model as the ceiling. This indicates that the proposed approach may be applied together with other approaches where updates to parameterized knowledge inside models can be done, therefore bringing the ceiling performance above that of the pretrained model. 

The critical token loosely defined in this work only includes entity names, locations and so on, which are the important tokens for factual QA tasks. However more rigorous definitions of it are needed to understand the issue of hallucination from an evaluator's perspective. In other words, this work only gives a very narrow and practical definition of critical token, but it is important to find the hallucinated generations, explicate the reason behind the judgments of hallucination, and examine how those judgments manifest in surface forms in a wide range of tasks for a more general definition of critical token.

Approaches in the work assume that alignment harms factuality of the model, leading to lower performance on factual QA tasks. This can be observed in experiments presented, but newer finetuning algorithms may alleviate or solve this issue, leading to finetuned models having the same or higher factuality. In this case, the proposed approaches will not be useful without careful redesign of different components.

\bibliography{custom}

\appendix

\section{Sampling for the baselines}

\begin{table}[h]
\centering
% \small
\begin{tabular}{lll}
\toprule
Baseline & TQA  & NQ \\
\midrule
CD & 64.19 \textcolor{red}{$_{-4.07}$} & 34.17 \textcolor{red}{$_{-2.80}$} \\
ITI  & 38.46 \textcolor{red}{$_{-10.32}$} & 14.64 \textcolor{red}{$_{-8.08}$} \\
DoLa & 68.56 \textcolor{red}{$_{-1.51}$} & 36.61 \textcolor{red}{$_{-0.67}$} \\
ICD  & 64.16 \textcolor{red}{$_{-4.90}$} & 34.67 \textcolor{red}{$_{-2.71}$} \\
\bottomrule
\end{tabular}

\caption{Experiment results of Llama 2 7B based strategies on the QA datasets with random sampling at temperature 1.0. The red numbers indicate the difference to the greedy decoding performance reported in Table \ref{tab:main_result}}
\label{tab:sampling}
\end{table}

Since the previous baselines usually employ greedy decoding as its decoding strategy, this experiment examines how their performance changes when sampling is used. This is to conduct a comparison for Model CDS's proposed advantage of being robust against random sampling. Table \ref{tab:sampling} shows the performance of the baselines with sampling on the QA datasets. All of their performances decrease substantially, with the largest drop seen in ITI. DoLa, however, is robust to sampling with only slight performance decrease, which may be caused by the fact that the final distributions are very spiky after contrasting. ITI's overall weak performance in either the greedy setting and the sampling setting may be due to that the model tends to give a general description of an entity within the question without directly answering the question. 

\section{System prompts}

The system prompt used in the TriviaQA and NaturalQuestions experiments is 

\texttt{You are a helpful, respectful and honest assistant. Always answer as helpfully as possible.}

for Llama 2 models. Note that this is shorter and less complicated than the original system prompt from Llama 2, because the original prompt causes the models to be overly conservative and refuse a majority of the questions in the datasets. For Mistral models, there is no system prompt.

The system prompt used in the \textsc{FActScore} experiments is 

\texttt{You are a helpful, respectful and honest assistant. Always answer as helpfully as possible, while being safe.  Your answers should not include any harmful, unethical, racist, sexist, toxic, dangerous, or illegal content. Please ensure that your responses are socially unbiased and positive in nature.}

This follows \citet{zhang2023alleviating} in order to compare to previous results. This prompt is also simpler than the original one, although it is slightly more complicated than the one above.

\section{Inference efficiency}

Model CDS has additional inference overhead compared to vanilla sampling. The overhead includes inference cost from the classifier and from the pretrained model, which depends on the sizes of the models in the whole setup. For example, if the models in Model CDS are all 13B models, on top of the normal inference cost $nk$, we expect additional classifier inference cost $nk$ and pretrained inference cost $0.1nk+t$, assuming 10\% of the tokens need regeneration. $n$ is the number of tokens, $k$ is inference cost for one token, and $t$ represents the inference cost of processing the context for the pretrained model. Inference efficiency can be gained from using small aligned models paired with large pretrained models, and using small classifier models. Compared to contrastive decoding with the same model setup, Model CDS requires more computation due to the assumed large classifier. However in the case of 70B model setup, Model CDS is more efficient than contrastive decoding, because Model CDS does not require both models to generate the whole response.

\end{document}